# Knowledge database development by large language models for countermeasures against viruses and marine toxins

Hung N. Do[1], Jessica Z. Kubicek-Sutherland[2], and S. Gnanakaran[1,*]

[1]Theoretical Biology and Biophysics Group, Theoretical Division; and [2]Physical Chemistry and Applied Spectroscopy Group, Chemistry Division, Los Alamos National Laboratory, Los Alamos, New Mexico 87545, USA

[*]Correspondence: gnana@lanl.gov

**Abstract**

Access to the most up-to-date information on medical countermeasures is important for the research and development of effective treatments for viruses and marine toxins. However, there is a lack of comprehensive databases that curate data on viruses and marine toxins, making decisions on medical countermeasures slow and difficult. In this work, we employ two large language models (LLMs) of ChatGPT and Grok to design two comprehensive databases of therapeutic countermeasures for five viruses of Lassa, Marburg, Ebola, Nipah, and Venezuelan equine encephalitis, as well as marine toxins. With high-level human-provided inputs, the two LLMs identify public databases containing data on the five viruses and marine toxins, collect relevant information from these databases and the literature, iteratively cross-validate the collected information, and design interactive webpages for easy access to the curated, comprehensive databases. Notably, the ChatGPT LLM is employed to design agentic AI workflows (consisting of two AI agents for research and decision-making) to rank countermeasures for viruses and marine toxins in the databases. Together, our work explores the potential of LLMs as a scalable, updatable approach for building comprehensive knowledge databases and supporting evidence-based decision-making.

***Keywords:*** *database, virus, marine toxin, therapeutic, countermeasure, large language model, agentic AI*

**Introduction**

Viruses and marine toxins are two of the many biothreats that pose significant harm to our society. For many viruses, information regarding their therapeutic treatments is spread across several publications[1-5], clinical trial registries ([https://clinicaltrials.gov/](https://clinicaltrials.gov/), [https://www.who.int/tools/clinical-trials-registry-platform/the-ictrp-search-portal](https://www.who.int/tools/clinical-trials-registry-platform/the-ictrp-search-portal), [https://www.who.int/tools/clinical-trials-registry-platform](https://www.who.int/tools/clinical-trials-registry-platform), [https://www.isrctn.com/](https://www.isrctn.com/), [https://www.clinicaltrialsregister.eu/](https://www.clinicaltrialsregister.eu/)), regulatory databases ([https://www.fda.gov/drugs/development-approval-process-drugs/drug-approvals-and-databases](https://www.fda.gov/drugs/development-approval-process-drugs/drug-approvals-and-databases), [https://www.fda.gov/drugs/drug-approvals-and-databases/drugsfda-data-files](https://www.fda.gov/drugs/drug-approvals-and-databases/drugsfda-data-files), [https://www.ema.europa.eu/en/medicines](https://www.ema.europa.eu/en/medicines), [https://products.mhra.gov.uk/](https://products.mhra.gov.uk/), [https://www.canada.ca/en/health-canada/services/drugs-health-products/drug-products/drug-product-database.html](https://www.canada.ca/en/health-canada/services/drugs-health-products/drug-products/drug-product-database.html)), and chemistry and bioactivity repositories[6,7] ([https://www.ebi.ac.uk/chembl/](https://www.ebi.ac.uk/chembl/), [https://go.drugbank.com/](https://go.drugbank.com/), [https://www.guidetopharmacology.org/](https://www.guidetopharmacology.org/)). The result is inconsistency in nomenclature and reporting of outcomes, as well as the mixing of animal models and human data. The same issues apply for the domain of marine toxins, where data are scattered across several publications[8-11], toxin databases[12,13] ([https://toxins.hais.ioc-unesco.org/](https://toxins.hais.ioc-unesco.org/), [https://haedat.iode.org/](https://haedat.iode.org/), [https://data.hais.ioc-unesco.org/](https://data.hais.ioc-unesco.org/), [https://marinespecies.org/hab/](https://marinespecies.org/hab/), [https://marinlit.rsc.org/](https://marinlit.rsc.org/)), and algal bloom surveillance portals ([https://www.cdc.gov/ohhabs/data/index.html](https://www.cdc.gov/ohhabs/data/index.html), [https://habsos.noaa.gov/](https://habsos.noaa.gov/), [https://coastalscience.noaa.gov/project/harmful-algal-bloom-hab-forecasting/](https://coastalscience.noaa.gov/project/harmful-algal-bloom-hab-forecasting/), [https://oceanservice.noaa.gov/hazards/hab/](https://oceanservice.noaa.gov/hazards/hab/), [https://hab.whoi.edu/maps/regions-world-distribution/](https://hab.whoi.edu/maps/regions-world-distribution/)).

Large language models (LLMs) serve as practical tools for extracting and summarizing information from broad public sources. However, they are known to introduce hallucinated facts as well as provide incomplete and inaccurate information[14,15]. While recent developments in LLMs (including ChatGPT 5.2, [https://openai.com/index/introducing-gpt-5-2/](https://openai.com/index/introducing-gpt-5-2/)) have been claimed to reduce the described issues[16-18], high-level human supervision is still required to ensure sufficient and accurate information collection by LLMs, especially for bio-surveillance and countermeasure assessments.

In this work, we described an approach employing LLMs to construct two comprehensive knowledge databases, one for viral therapeutic countermeasures and one for marine toxins, and deploy them as interactive webpages with high-level human inputs. Following human-provided prompts, the LLMs wrote the necessary codes to carry out the necessary tasks to construct the websites, including public database identification, data curation and verification, and website design (**Figure 1A**). Notably, the LLMs were employed to design online agentic AI workflows comprising two AI agents each to summarize and rank potential countermeasures for viruses and marine toxins (**Figure 1B**). Our work illustrated the potential of LLMs to function effectively, under human supervision, to accelerate several tasks associated with the creation of knowledge databases.

**Details**

We constructed two knowledge databases of viral therapeutic countermeasures (excluding vaccines) for the five viruses of Lassa (LASV), Marburg (MARV), Zaire ebolavirus (EBOV), Nipah (NiV), and Venezuelan equine encephalitis virus (VEEV), as well as of marine toxins. The two databases were constructed entirely by LLMs, with high-level human input and verification. The overall workflow to develop the interactive webpages included searches of public databases, data curation and verification using retrieval-augmented generation (RAG) using public databases and literature, webpage designs, and incorporation of agentic AI (with one 'researcher' and one 'decision-maker' agent) into the designed webpages for ranking of countermeasures (**Figure 1**).

***Data curation by LLMs for the web portal of therapeutic countermeasures against viruses***

We employed primarily "ChatGPT 5.2 Thinking" of Los Alamos National Laboratory ChatGPT Enterprise as well as "Grok 4.1 Thinking" to build the webpage for viral therapeutic (TRx) countermeasures by performing a deep search and curating data related to the therapeutic countermeasures for five viruses, including LASV, MARV, EBOV, NiV, and VEEV (**Figure 1A**). Here, ChatGPT was employed as the primary LLM to collecting the relevant information for the viruses, while Grok was employed for cross-validation of the information collected by ChatGPT. We defined the primary purpose of cross-validation to be ensuring the information collected by the two LLMs was consistent with each other, and no relevant information was overlooked during the data collection. The collected information was verified iteratively using ChatGPT first and then by another LLM (Grok).

First, we queried ChatGPT to retrieve reliable public databases containing information regarding the therapeutic countermeasures (excluding vaccines) for viruses. Here, we considered therapeutic countermeasures that included prophylaxis, antibodies, antivirals, host-directed agents, repurposed drugs, and supportive care. The LLM's output was included in **Main Data 1**. Afterwards, using the RAG method, we prompted the LLM to consult databases listed in **Main Data 1** and publications in the National Library of Medicine (PubMed, https://pubmed.ncbi.nlm.nih.gov/) and Europe PMC (https://europepmc.org/) for relevant information on viral therapeutic countermeasures (**Figure 1A**). Our interest has been in information on native survival rates and viremia for the five viruses, along with their potential therapeutic countermeasures. We grouped therapeutic countermeasures into three categories: pathogen-targeted treatments, host-targeted treatments, and combinatorial strategies.

For each therapeutic countermeasure, information on survival rates and viremia post-treatment (if available), mechanisms of action, treatment types, dosages, and effectiveness was collected, along with the associated references. We considered both in vivo and in vitro data as well as data from both animal models and human clinical trials. Furthermore, the LLMs were allowed to collect data from both preprint and peer-reviewed manuscripts. We represented survival rates and viremia over days post-infection (dpi) as plots and only queried the LLMs to point to publications containing the relevant images for manual collection to ensure the best image quality. Meanwhile, we queried the LLMs to automatically extract and summarize information on the mechanisms of action, treatment types, dosages, and effectiveness of the therapeutic countermeasures for inclusion on the TRx webpage. The therapeutic countermeasures collected for the five viruses of Lassa, Marburg, Ebola, Nipah, and Venezuelan equine encephalitis were included in **Main Data 2**. All collected information was stored in a JSON file for incorporation into the TRx webpage. The information in the JSON file has been updated monthly since November 2025 using LLMs to ensure the TRx webpage contains the most relevant information on therapeutic countermeasures for the five viruses.

***Data curation by LLM for the web portal of therapeutic countermeasures against marine toxins***

We used the same LLMs to perform deep search and curate data on marine toxins (**Figure 1A**). Here, Grok was employed as the primary LLM to collect information on marine toxins, whereas ChatGPT was employed for cross-validation due to safety guardrails considerations. Consequently, the collected information was verified iteratively first with Grok and then with another LLM (ChatGPT).

We also queried ChatGPT to retrieve reliable public databases containing information regarding marine toxins. The LLM's output was included in **Main Data 3**. Then, we again employed the RAG method and prompted Grok to review databases listed in **Main Data 3** and publications in the National Library of Medicine (PubMed, https://pubmed.ncbi.nlm.nih.gov/) and Europe PMC (https://europepmc.org/) to identify all relevant information on marine toxins. The LLMs categorized the marine toxins into 16 different families, including conotoxins, paralytic shellfish toxins (saxitoxins), tetrodotoxins, brevetoxins, ciguatoxins, domoic acids, okadaic acids and dinophysistoxins, pectenotoxins, azaspiracids, yessotoxins, palytoxins, maitotoxins, cyclic imines, gambierols, karlotoxins, and amphidinols (**Main Data 4**). For each toxin, we queried the LLMs to collect information on their classes, along with concerns about bioweapon use (if available), to serve as the main tabs for each toxin family (**Figure 1A**). We limited the biothreat concerns to risk communication only to avoid malicious use. For each class of the toxin, we prompted the LLMs to collect and summarize information on their source organisms, host molecular targets, analogues, toxicity, exposure syndromes, potential countermeasures, and associated references for inclusion on the marine toxin webpage (**Main Data 4**). Similar to TRx, the collected information by the LLMs was stored in a JSON file and has been updated monthly since November 2025 by the LLMs to ensure the marine toxin webpage contains the most up-to-date information regarding the marine toxins.

***Design of the web portal of viral therapeutic countermeasures and marine toxins by LLMs***

We provided the JSON files containing the collected data on viruses and marine toxins to "ChatGPT 5.1 and 5.2 Thinking" to design their webpages (**Figure 1A**). The LLMs took in our prompts and generated *python* scripts based on FastAPI (https://fastapi.tiangolo.com/) with Pydantic incorporated (https://docs.pydantic.dev/latest/) to design HTML files containing the webpages for viral therapeutic countermeasures and marine toxins (**Supplementary Data 1** and **2**). The webpage snapshots of the web portal are shown in **Figures 2** and **3**.

The webpage for viral therapeutic (TRx) countermeasures, titled "Viral Therapeutic (TRx) Countermeasure Explorer", was designed to include four hierarchical tab levels of "Viruses", "Countermeasure Categories, "Treatments," and "Treatment Details". Initially, only the "Viruses" tab, including the five virus names of "Lassa virus (LASV)", "Marburg virus (MARV)", "Zaire ebolavirus (EBOV)", "Nipah virus (NiV)", and "Venezuelan equine encephalitis virus (VEEV)", would be available to users (**Figure 2A**). Clicking on one of the viruses revealed the figure describing the native survival rates and viremia post-infection with the virus, along with the associated references for the figure, as well as the "Countermeasure Categories" panel with three clickable tabs of "Pathogen-Targeted Treatment", "Host-Targeted Treatment", and "Combinatorial Strategies". Clicking on one of the countermeasure categories revealed the treatments included in that category, as specified in **Main Data 2**. Clicking on a treatment revealed details on its mechanisms of action, treatment type, dosage, effectiveness, survival rates, and viremia post-treatment (if available), along with the associated references.

The web portal for marine toxins, titled "Marine Toxin Countermeasure Explorer", was designed to include three hierarchical levels. Initially, only a central circle labeled "Marine Toxins" would be visible to users (**Figure 3A**). Clicking on the circle revealed the 16 marine toxin families, including "Conotoxins, CTX", "Paralytic Shellfish Toxins, PST (Saxitoxins)", "Tetrodotoxin, TTX", "Brevetoxins, NSP", "Ciguatoxins, CFP", "Domoic Acid, ASP", "DSP Toxins (Okadaic acid and Dinophysistoxins", "Pectenotoxins, PTX", "Azaspiracids, AZA", "Yessotoxins, YTX", "Palytoxin, PLTX", "Maitotoxins, MTX", "Cyclic Imines (SPX, GYM, PnTX)", "Gambierol (GAM)", "Karlotoxins (KmTX), and "Amphidinols (AM)" presented as circles branching out from the central circle (**Figure 3A**). Clicking on the circle of one marine toxin family revealed one panel containing the "Biothreat Concerns" tab (if information was available) and tabs of the classes for the marine toxin family (as detailed in **Main Data 4**). Clicking on one of the classes revealed tabs containing information on its "Source Organisms", "Host Molecular Targets", "Analogues", "Exposure Syndromes", "Countermeasures", "Toxicity", and associated references (**Figure 3A**).

*Implementation of AI agents to rank countermeasures against viruses and marine toxins*

We prompted ChatGPT 5.2 Thinking to design agentic AI workflows to be incorporated into webpages that rank countermeasures for the five viruses and marine toxins (**Figure 1A-B**). The AI agents were designed based on the OpenAI GPT-5.2 model (https://openai.com/api/). Each agentic AI workflow was designed to consist of two AI agents: a 'researcher' and a 'decision-maker' (**Figure 1B**). The 'researcher' agent reviewed the collected information on the webpage and may collect additional information from reliable internet sources related to countermeasures to provide to the 'decision-maker' agent (**Figure 1B**). Afterwards, the 'decision-maker' agent evaluated the information provided by the 'researcher' agent and ranked the countermeasures for the virus or marine toxin based on their extents of mortality reductions (**Figure 1B**). The output provided by the 'decision-maker' agent included the ranking of the countermeasures and benefits in reducing mortality for the countermeasures, evidence strengths, rationales and constraints used in the ranking, along with the sources used in the ranking (**Figures 2B** and **3B**). The agentic AI workflows were built using LLM-generated *python* scripts, with constraints in place so that users could only ask questions related to the ranking of the countermeasures.

**Conclusion**

We constructed two comprehensive knowledge databases of therapeutic countermeasures against five viruses and 16 different marine toxin families using LLMs (ChatGPT and Grok), entirely based on high-level human inputs and verifications. The employment of two different LLMs allowed for primary curation plus cross-validation of data, ensuring no information was inaccurate or missing during data collection. Through human-provided inputs, the LLMs handled every single task involved in the designing of the two interactive knowledge databases, from data curation and verification, monthly data updates, webpage designs by generating relevant *python* scripts, and designing OpenAI GPT-5.2-based agentic AI (with one 'researcher' and one 'decision-maker' agent) to be incorporated into the designed webpages for the purpose of ranking countermeasures. We expect these two knowledge databases to serve as valuable research tools for developing medical countermeasures. Our work explored the potential of LLMs for collecting data and generating relevant codes to accomplish specified tasks, including building comprehensive databases and decision-making workflows, with high-level human-provided inputs. We demonstrated the abilities of LLMs not only as standalone applications but also as accelerating, auditable, and updatable workflows. Given our findings from this work, we expect the LLMs to be adopted to accelerate many other applications, including the development of other ML models for complicated tasks.

**Data Availability Statement**

Data supporting the findings of this study are included in the article and Supplementary Information files.

**Code Availability Statement**

Custom codes related to the curation and verification of data, the design of interactive webpages, and the design of agentic AI platforms were generated by "LANL ChatGPT Enterprise 5.1 and 5.2 Thinking" and "Grok 4.1 Thinking". The custom codes may be made available upon approval from funding sponsors.

**Conflict of Interest Statement**

The authors declare no conflict of interest.

**Acknowledgements**

This work was supported by the Defense Threat Reduction Agency under Grant no. HDTRA1551890. The views expressed in this article are those of the authors and do not reflect the official policy or position of the U.S. Department of Defense or the U.S. Government. We would like to thank the Los Alamos National Laboratory Institutional Computing and AI portal for their computing resources. The authors thank Dr. Julie Barbaras and Lieutenant Colonel Tiffany Nguyen for their guidance and support.

**Author Contributions**

H.N.D. performed research, analyzed data, and wrote the manuscript. J.Z.K. acquired funding, supervised research, analyzed data, and wrote the manuscript. S.G. supervised research, analyzed data, and wrote the manuscript. All authors contributed to the final version of the manuscript.

**References**

1. Eberhardt, K. A. *et al.* Ribavirin for the treatment of Lassa fever: A systematic review and meta-analysis *Int J Infect Dis* **87**, 15-20 (2019). https://doi.org/10.1016/j.ijid.2019.07.015
2. Salam, A. P. *et al.* Time to reconsider the role of ribavirin in Lassa fever. *PLOS Neglected Tropical Diseases* **15**, e0009522 (2021). https://doi.org/10.1371/journal.pntd.0009522
3. Chan, X. H. S. *et al.* Therapeutics for Nipah virus disease: a systematic review to support prioritisation of drug candidates for clinical trials *Lancet Microbe* **6**, 101002 (2025). https://doi.org/10.1016/j.lanmic.2024.101002
4. Bossart, K. N. *et al.* A neutralizing human monoclonal antibody protects against lethal disease in a new ferret model of acute nipah virus infection *PLOS Pathogens* **5**, e1000642 (2009). https://doi.org/10.1371/journal.ppat.1000642
5. Ogorek, T. J. & Golden, J. E. Advances in the Development of Small Molecule Antivirals against Equine Encephalitic Viruses *Viruses* **15**, 413 (2023). https://doi.org/10.3390/v15020413
6. Wang, Y. *et al.* An overview of the PubChem BioAssay resource. *Nucleic Acids Research* **38**, D255-D266 (2010). https://doi.org/10.1093/nar/gkp965
7. Liu, T. *et al.* BindingDB in 2024: a FAIR knowledgebase of protein-small molecule binding data. *Nucleic Acids Research* **53**, D1633-D1644 (2024). https://doi.org/10.1093/nar/gkae1075
8. Yuan, K.-K., Li, H.-Y. & Yang, W.-D. Marine Algal Toxins and Public Health: Insights from Shellfish and Fish, the Main Biological Vectors. *Marine Drugs* **22**, 510 (2024). https://doi.org/10.3390/md22110510
9. Louzao, M. C. *et al.* Current Trends and New Challenges in Marine Phycotoxins. *Marine Drugs* **20**, 198 (2022). https://doi.org/10.3390/md20030198
10. Farabegoli, F., Blanco, L., Rodriguez, L. P., Vieites, J. M. & Cabado, A. G. Phycotoxins in Marine Shellfish: Origin, Occurrence and Effects on Humans. *Marine Drugs* **16**, 188 (2018). https://doi.org/10.3390/md16060188
11. Do, H. N., Kubicek-Sutherland, J. Z. & Gnanakaran, S. Prediction of Specificity of α-Conotoxins to Subtypes of Human Nicotinic Acetylcholine Receptors with Semi-Supervised Machine Learning. *ACS Chemical Neuroscience* **16**, 2196-2207 (2025). https://doi.org/10.1021/acschemneuro.4c00760
12. Kaas, Q., Westermann, J.-C. & Craik, D. J. Conopeptide characterization and classifications: An analysis using ConoServer. *Toxicon* **55**, 1491-1509 (2010). https://doi.org/https://doi.org/10.1016/j.toxicon.2010.03.002

13  Kaas, Q., Yu, R., Jin, A. H., Dutertre, S. & Craik, D. J. ConoServer: updated content, knowledge, and discovery tools in the conopeptide database. *Nucleic Acids Research* **40**, D325-330 (2012).
14  Kalai, A. T., Nachum, O., Vempala, S. S. & Zhang, E. Why Language Models Hallucinate. *arXiv preprint* (2025). https://doi.org/10.48550/arXiv.2509.04664
15  Jiang, C. *et al*. On Large Language Models' Hallucination with Regard to Known Facts. *arXiv preprint* (2024). https://doi.org/10.48550/arXiv.2403.20009
16  Béchard, P. & Ayala, O. M. Reducing hallucination in structured outputs via Retrieval-Augmented Generation. *arXiv preprint* (2024). https://doi.org/10.48550/arXiv.2404.08189
17  Lewis, P. *et al*. Retrieval-Augmented Generation for Knowledge-Intensive NLP Tasks. *arXiv preprint* (2020). https://doi.org/10.48550/arXiv.2005.11401
18  Zhang, W. & Zhang, J. Hallucination Mitigation for Retrieval-Augmented Large Language Models: A Review. *Mathematics* **13**, 856 (2025). https://doi.org/10.3390/math13050856

**Figures**

**Figure 1. Workflow for developments of the web portal by LLMs for therapeutic countermeasures against viruses and marine toxins (A)** Two different LLMs (ChatGPT and Grok) were employed for curation and iterative verification of data, which were provided to ChatGPT for the design of the webpages. ChatGPT was also used to design agentic AI workflows to be incorporated into the webpages for ranking of the countermeasures based on the collected information. High-level human inputs were provided to the LLMs to carry out the tasks. **(B)** Scheme of the agentic AI workflows that were designed to rank countermeasures based on the collected information. A 'researcher' agent was included to extract and summarize relevant information regarding the countermeasures for the biothreat of interest, which was then provided to a 'decision-maker' agent for ranking.

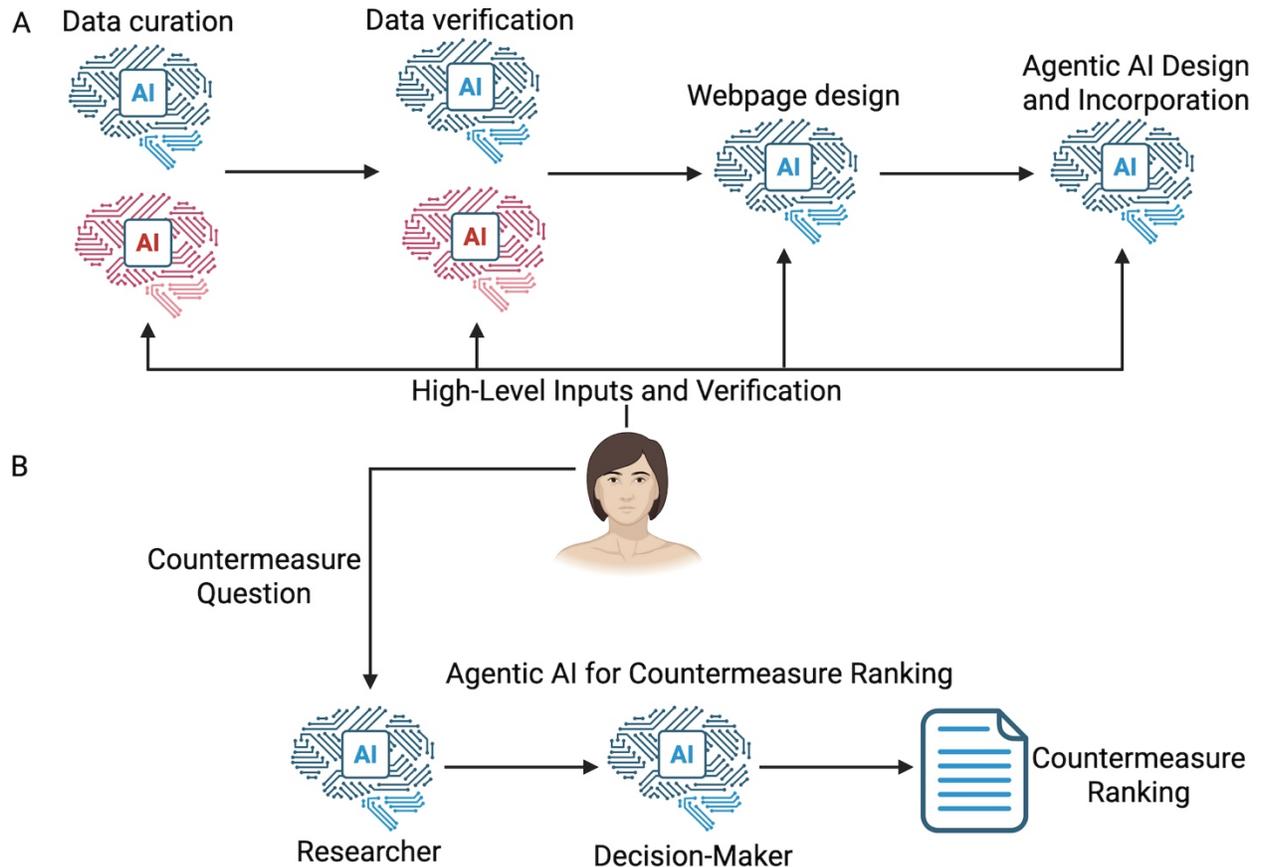

**Figure 2. Interactive web portal for the comprehensive database of viral therapeutic countermeasures as designed by LLMs. (A)** Representative snapshot of the interactive webpage for the comprehensive database of viral therapeutic countermeasures. **(B)** Representative snapshot of a ranking for countermeasures of Nipah virus as provided by the agentic AI.

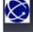

**Figure 3. Interactive webpage for the comprehensive database of marine toxins as designed by LLMs. (A)** Representative snapshot of the interactive webpage for the comprehensive database of marine toxin. **(B)** Representative snapshot of a ranking for countermeasures of conotoxins as provided by the agentic AI.

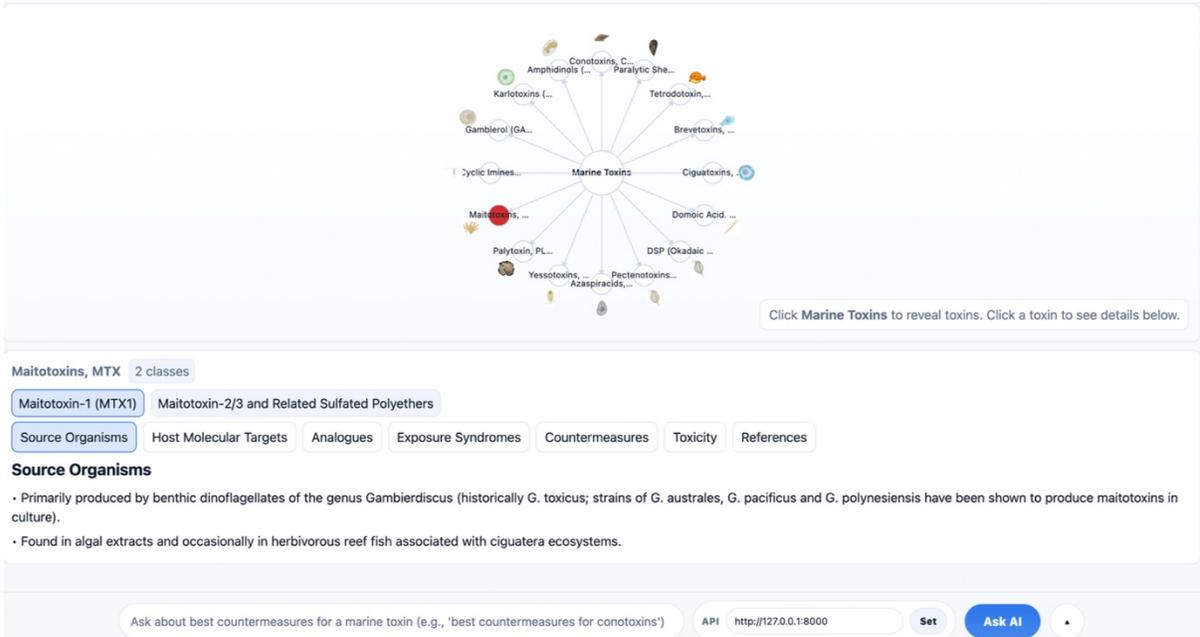

**Main Data 1. Public databases related to the therapeutic countermeasures for viruses.** Widely used public registries relevant to antiviral and antibody countermeasures were included in the list.

Chemistry / Bioactivity / Drug-Target
- PubChem (NCBI/NIH)
- ChEMBL (EMBL-EBI)
- DrugBank
- Open Targets Platform
- Drug Repurposing Hub (Broad Institute)
- BindingDB
- IUPHAR / Guide to Pharmacology
- Protein Data Bank (PDB)

Immunology / Epitopes / Antibodies
- Immune Epitope Database (IEDB)
- SAbDab (Structural Antibody Database)
- Thera-SAbDab (Therapeutic Structural Antibody Database)

Pathogen / Genomics / Evolution
- BV-BRC (Bacterial and Viral Bioinformatics Resource Center)
- NIAID Bioinformatics Resource Centers (BRCs) program directory
- GISAID
- Nextstrain

Clinical Trial Registries
- ClinicalTrials.gov
- WHO ICTRP (International Clinical Trials Registry Platform)
- EU Clinical Trials Register
- ISRCTN
- ANZCTR (Australian New Zealand Clinical Trials Registry)
- ChiCTR (Chinese Clinical Trial Registry)
- CTRI (Clinical Trials Registry – India)
- Pan African Clinical Trials Registry (PACTR)

Regulatory Product Databases
- Drugs@FDA
- EMA EPAR (European Public Assessment Reports)
- Health Canada Drug Product Database (DPD)
- MHRA products information (UK)
- TGA ARTG (Australia)
- PMDA approvals information (Japan)

Pathogen-Specific
- Stanford HIV Drug Resistance Database (HIVDB)

**Main Data 2. Therapeutic countermeasures collected for the five viruses in the TRx webpage by the LLMs.**

Lassa virus (LASV)
- Pathogen-Targeted Treatments: Arevirumab-3, Favipiravir, LHF-535, Ribavirin (IV), and ST-193
- Host-Targeted Treatments: 25-Hydroxycholesterol, PF-429242, PRTX007
- Combinatorial Strategies: Arevirumab-3 + Favipiravir, Dexamethasone + Ribavirin, LHF-535 + Favipiravir, Ribavirin + Favipiravir, Ribavirin + LHF-535

Marburg virus (MARV)
- Pathogen-Targeted Treatments: AVI-7288 phosphorodiamidate morpholino oligomer, Favipiravir, Galidesivir, MBP091, MR186-YTE monoclonal antibody, MR191-N monoclonal antibody, NP-718m-LNP, Remdesivir, siRNA-LNP therapy
- Host-Targeted Treatments: Infection prevention & control, supportive critical care
- Combinatorial Strategies: MR186-YTE + remdesivir

Zaire ebolavirus (EBOV)
- Pathogen-Targeted Treatments: AVI-7537, Ebanga, Favipiravir, Inmazeb, Remdesivir, TKM-130803, ZMapp
- Host-Targeted Treatments: Convalescent plasma, optimized supportive/critical care bundle
- Combinatorial Strategies: Convalescent plasma + supportive care, monoclonal antibody + optimized supportive care

Nipah virus (NiV)
- Pathogen-Targeted Treatments: Defective-interfering particles, engineered soluble ephrin-B2 decoy receptor, Favipiravir, fusion-inhibitory lipopeptides, HENV-103 + HENV-117, hu1F5 / MBP1F5, Lumicitabine / ALS-8112, m102.4, Remdesivir, Ribavirin
- Host-Targeted Treatments: Infection prevention & control, public-health measures, supportive critical care
- Combinatorial Strategies: Early mAb + remdesivir, PEP for high-risk contacts: m102.4 + contact tracing

Venezuelan equine encephalitis virus (VEEV)
- Pathogen-Targeted Treatments: BDGR-164, BDGR-4, BDGR-49, c1A3B-7, Hu-1A4A-1-YTE, ML336 and derivatives, neutralizing monoclonal antibodies targeting E2, VEEV nsP2 protease inhibitors, $\beta$-D-N4-hydroxycytidine / Molnupiravir
- Host-Targeted Treatments: general infection prevention & control (IPC) in healthcare, neurocritical care & supportive management for viral encephalitis, type I interferon therapy

**Main Data 3. Public databases related to marine toxins.**

Global / Core HAB & Phycotoxin Portals
- HAIS (Harmful Algal Information System) – IOC-UNESCO
- HAEDAT (Harmful Algal Event Database) – IOC-UNESCO + ICES + PICES
- Taxonomic Reference List of Harmful Microalgae – IOC-UNESCO / WoRMS
- OBIS HAB datasets – OBIS
- ICES-IOC WGHABD – ICES area
- EURL for Monitoring of Marine Biotoxins – EU reference network

National / Regional HAB Monitoring
- NOAA NCCOS HAB Portal & Monitoring System – USA
- FWC / FWRI HAB Monitoring Database (Karenia brevis) – Florida, USA
- California HABMAP (marine) + CDPH Marine Biotoxin Program – California, USA
- Alaska HAB Network (AHAB) Data Portal – Alaska, USA
- SoundToxins – Washington, USA (Puget Sound)
- IFOP / Chile "Marea Roja" (Red Tide) Program – Chile
- Canada (DFO/CFIA contributions to HAEDAT) – Canada
- Australia National Compilations (IMOS / state programs)
- US EPA HAB Data (coasts & inland) – USA

Toxicology / Chemistry / Mechanism Databases
- EFSA OpenFoodTox – EU
- T3DB (Toxin & Toxin-Target Database)
- CompTox Chemicals Dashboard (US EPA)
- CTD (Comparative Toxicogenomics Database)
- PubChem (NCBI/NIH)
- CyanoMetDB (cyanobacterial metabolites / cyanotoxins)

Protein / Peptide Toxins
- UniProt Tox-Prot
- ConoServer

Case / Exposure Surveillance
- Ciguatera Map & Reporting System (Ciguawatch)
- Water Quality Portal (WQP)
- WQX (Water Quality Exchange)

**Main Data 4. Classes found for each of the 16 marine toxin families by the LLMs.**

| |
|---|
| Conotoxins (10 classes) |
|     - $\alpha$-conotoxins, $\omega$-conotoxins, $\mu$-conotoxins, $\mu$O-conotoxins, $\kappa$-conotoxins, $\delta$-conotoxins, $\iota$-conotoxins, conantonkins, $\chi$-conopeptides, and $\rho$-conotoxins |
| Paralytic shellfish toxins (saxitoxins) (6 classes) |
|     - Carbamate toxins, decarbamoyl toxins, N-sulfocarbamoyl toxins, deoxydecarbamoyl toxins, benzoate PSTs, and M-toxins |
| Tetrodotoxin (5 classes) |
|     - Tetrodotoxin (TTX), tetrodotoxin (canonical), anhydro/epimeric analogues, deoxy analogues, and oxidized/other analogues |
| Brevetoxins (5 classes) |
|     - Brevetoxins (PbTx-1 to -10), Brevetoxin-B backbone analogs (PbTx-2 group), Brevetoxin-A backbone analogs (PbTx-1 group), conjugates & metabolites in shellfish/mammals, and Brevenal & site-5 antagonists (endogenous inhibitors) |
| Ciguatoxins (4 classes) |
|     - Pacific ciguatoxins, Caribbean ciguatoxins, Indian Ocean ciguatoxins, and algal precursors & fish metabolites |
| Domoic acid (4 classes) |
|     - Domoic acid, domoic acid (canonical), epi- and isodomoic acids, and seafood vectors & matrices |
| Okadaic acid and dinophysistoxins (5 classes) |
|     - Okadaic acid & DTXs, okadaic acid, dinophysistoxin-1, dinophysistoxin-2, and dinophysistoxin-3 |
| Pectenotoxins (3 classes) |
|     - Pectenotoxins (PTX), parent algal pectenotoxins (PTX1/2/6/11), and shellfish metabolites & esters (PTX2sa, 7-epi-PTX2sa, acyl esters) |
| Azaspiracids (3 classes) |
|     - Azaspiracids (AZA), parent algal azaspiracids (AZA-1/2/3), additional algal analogues (AZA-38/-39, AZA-59) |
| Yessotoxins (3 classes) |
|     - Yessotoxins (YTX), parent algal yessotoxins (YTX / homoYTX / 45-OH-YTX), shellfish metabolites & analogues (carboxyYTX, norYTXs, 45,46,47-trinorYTX) |
| Palytoxin (2 classes) |
|     - Palytoxin (PLTX) & classic analogues, ostreopsis analogues (ovatoxins & ostreocins) |
| Maitotoxins (2 classes) |
|     - Maitotoxin-1 (MTX1), maitotoxin-2/3 and related sulfated polyethers |
| Cyclic imines (3 classes) |
|     - Spirolides, gymnodimines, pinnatoxins |
| Gambierol (1 class) |
|     - Gambierol |
| Karlotoxins (1 class) |
|     - Karlotoxins |
| Amphidinols (1 class) |
|     - Amphidinols |